\documentclass[10pt,twocolumn,letterpaper]{article}

\usepackage{cvpr}
\usepackage{times}
\usepackage{epsfig}
\usepackage{graphicx}
\usepackage{amsmath}
\usepackage{amssymb}
\usepackage{xspace}
\usepackage{booktabs}
\usepackage[vlined,linesnumbered,ruled,algo2e]{algorithm2e}
\usepackage[font=small]{caption}
\usepackage[linesnumbered,ruled,vlined]{algorithm2e}

\usepackage{multirow}
\usepackage{xcolor}

\usepackage[labelformat=simple]{subcaption}

\usepackage[pagebackref=true,breaklinks=true,letterpaper=true,colorlinks,bookmarks=false]{hyperref}

\usepackage[flushleft]{threeparttable}

\cvprfinalcopy 

\def\cvprPaperID{85} 

\ifcvprfinal\fi

\newcommand{\figLabel}{Figure\xspace}

\newcommand{\mysection}[1]{\vspace{3pt}\noindent\textbf{#1}}

\begin{document}



\title{When NAS Meets Trees: \\An Efficient Algorithm for Neural Architecture Search}

\author{
    Guocheng Qian $^{1}$ \quad 
	Xuanyang Zhang $^{2}$\quad 
	Guohao Li $^{1}$\quad 
    Chen Zhao $^{1}$\quad
	Yukang Chen $^{3}$ \\ 
	Xiangyu Zhang $^{2}$ \quad 
	Bernard Ghanem $^{1}$ \quad 
	Jian Sun $^{2}$ \\
   $^{1}$ King Abdullah University of Science and Technology (KAUST), Thuwal, Saudi Arabia\\
   $^{2}$ MEGVII Technology \quad
   $^{3}$ The Chinese University of Hong Kong\\
   {\tt\small \{guocheng.qian, bernard.ghanem\}@kaust.edu.sa}
}

\maketitle

\begin{abstract}
The key challenge in neural architecture search (NAS) is designing how to explore wisely in the huge search space. We propose a new NAS method called TNAS (NAS with trees), which improves search efficiency by exploring only a small number of architectures while also achieving a higher search accuracy. TNAS introduces an architecture tree and a binary operation tree, to factorize the search space and substantially reduce the exploration size. TNAS performs a modified bi-level Breadth-First Search in the proposed trees to discover a high-performance architecture. Impressively, TNAS finds the global optimal architecture on CIFAR-10 with test accuracy of 94.37\% in four GPU hours in NAS-Bench-201. The average test accuracy is 94.35\%, which outperforms the state-of-the-art. Code is available at: \url{https://github.com/guochengqian/TNAS}. 
\end{abstract}

\section{Introduction}
\label{sec:introduction}

Neural architecture search has spurred increasing interest in both academia and industry for its ability in finding high-performance neural network architectures with minimal human intervention. To achieve the most accurate NAS algorithm, one can explore all candidate architectures, training each one to convergence, and picking the best-performing architecture. However, this brute-force NAS is infeasible due to the enormous search space. Therefore, one of the key questions towards a successful NAS algorithm is: how to efficiently explore the search space?

One-shot NAS \cite{brock2017smash,pham2018efficient,bender2018understanding,liu2018darts} impressively improved the efficiency of NAS. One-shot NAS leverages a weight-sharing strategy and approximately trains only one network, called the \emph{supernet}, which subsumes all candidate architectures. Each candidate architecture directly inherits weights from the supernet without training.  Despite the efficiency of one-shot NAS algorithms, they incur architecture evaluation degradation, \ie the architecture performance evaluated using the weight-sharing is not correctthat, which leads to a degraded search accuracy \cite{Yu20Evaluting,li2020sgas}.


\begin{figure}[t]
\begin{minipage}[b]{0.55\columnwidth}
	    \begin{subfigure}{\textwidth}
	    	\includegraphics[width=\textwidth]{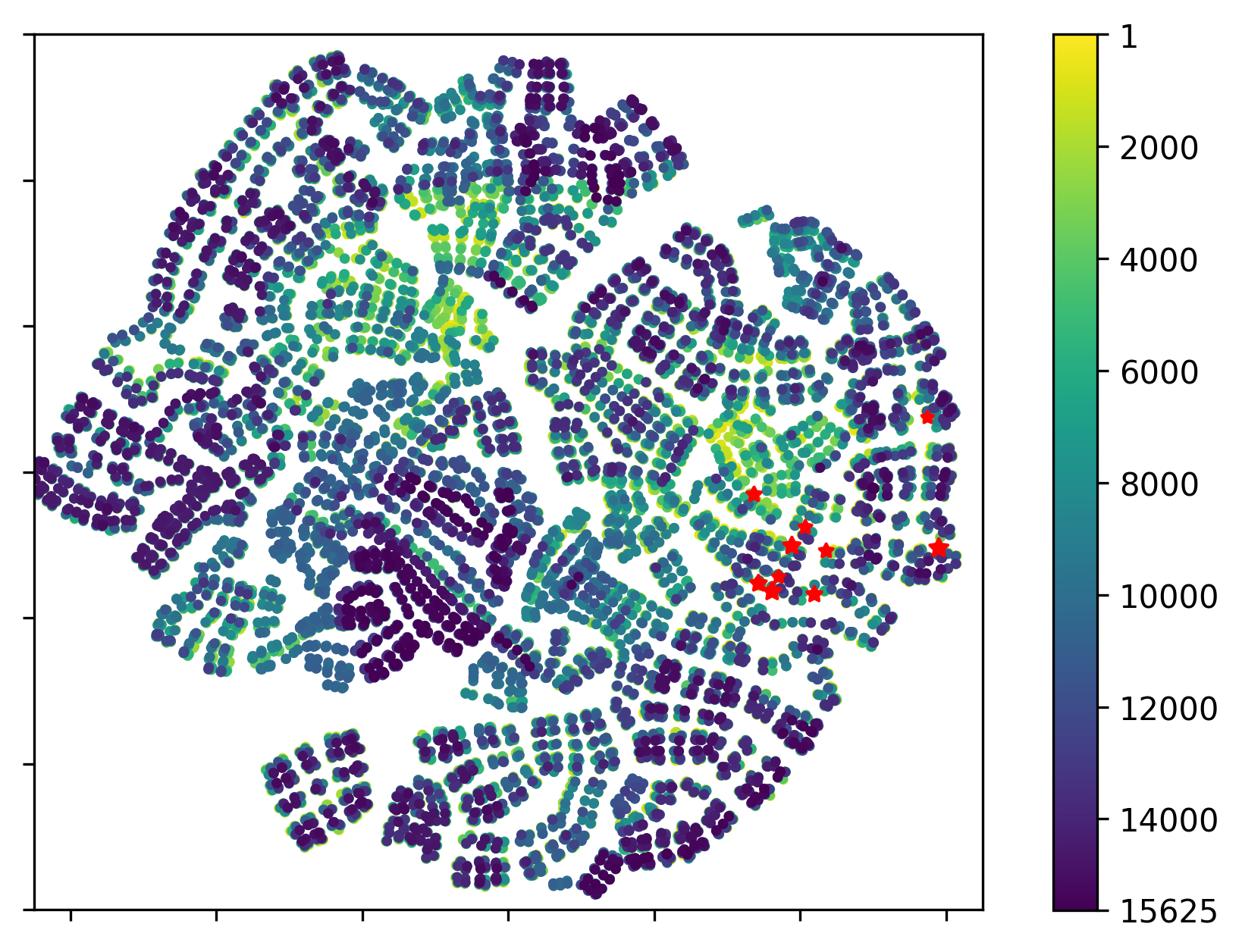}
	        \caption{The entire search space. Each \\dot represents an architecture.
}
	        \label{fig:pruning0}
	    \end{subfigure}%
\end{minipage}\hfill
\begin{minipage}[s]{0.44\columnwidth}
	    \begin{subfigure}{\textwidth}
	    	\includegraphics[width=\textwidth]{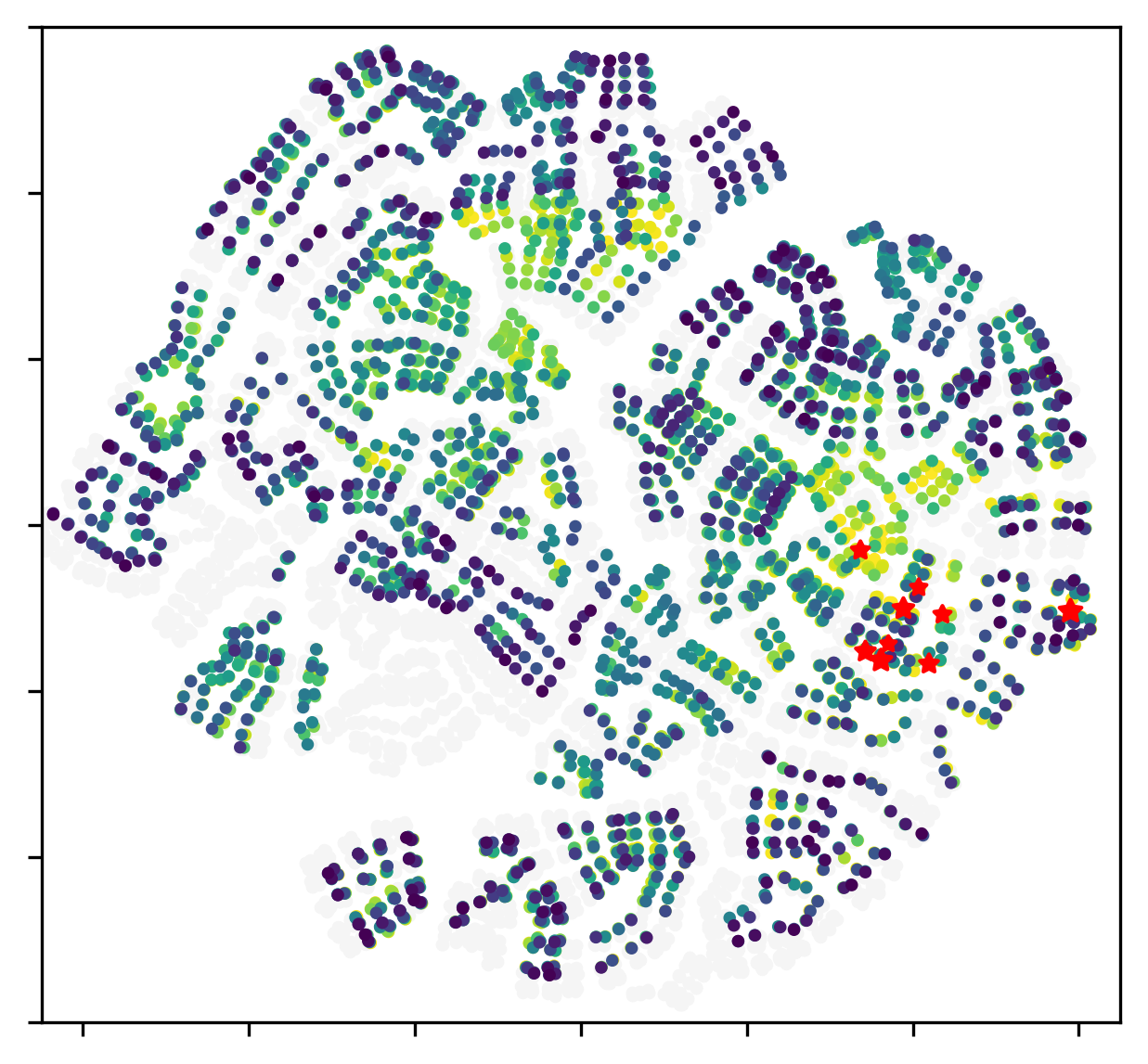}
	        \caption{The pruned search space after the first search stage. 
}
	        \label{fig:pruning1}
	    \end{subfigure}%
\end{minipage}
\begin{minipage}[b]{0.45\columnwidth}
	    \begin{subfigure}{\textwidth}
	    	\includegraphics[width=\textwidth]{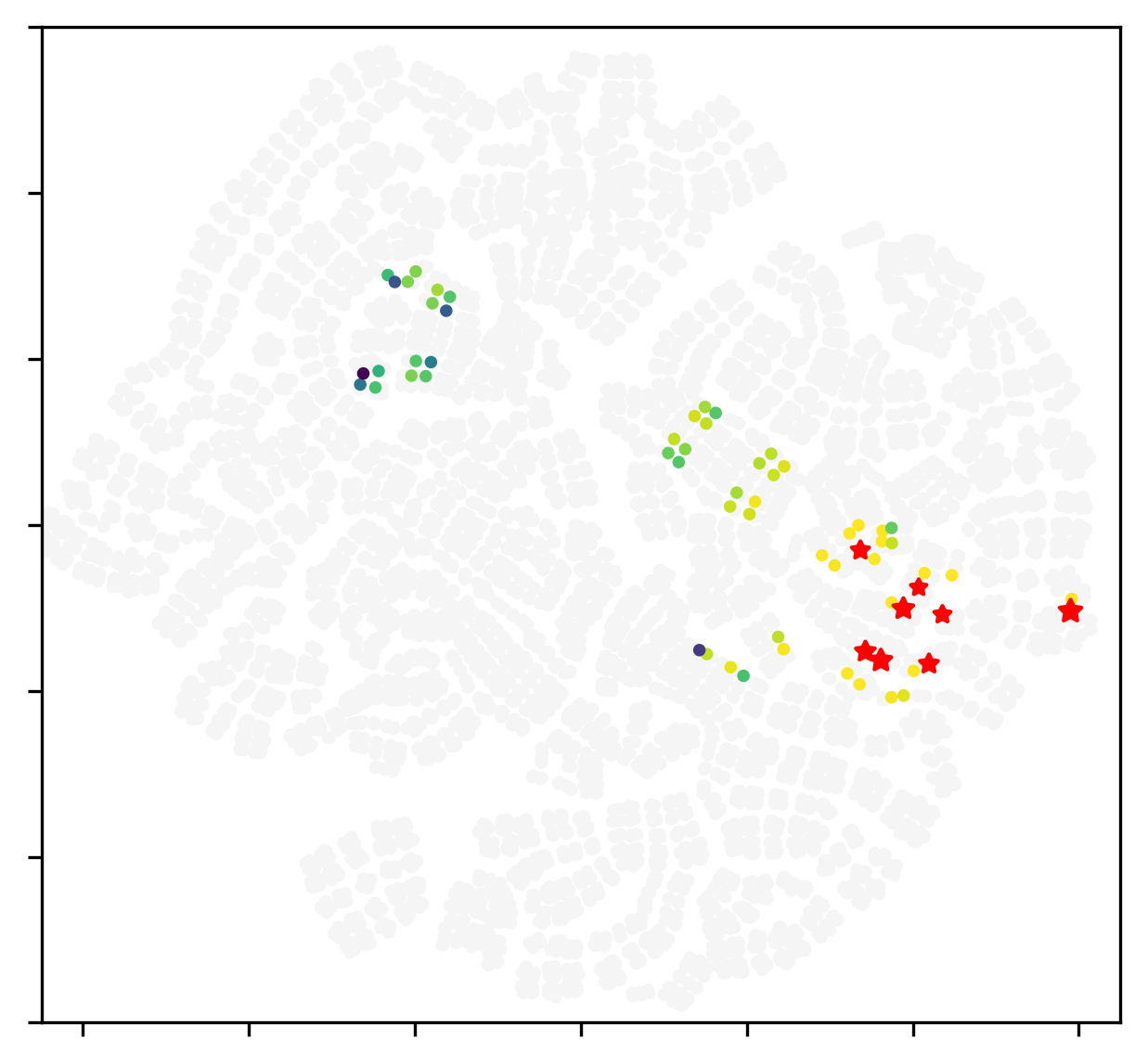}
	        \caption{The pruned search space after the second search stage. 
}
	        \label{fig:pruning2}
	    \end{subfigure}%
\end{minipage}\hfill
\begin{minipage}[s]{0.45\columnwidth}
	    \begin{subfigure}{\textwidth}
	    	\includegraphics[width=\textwidth]{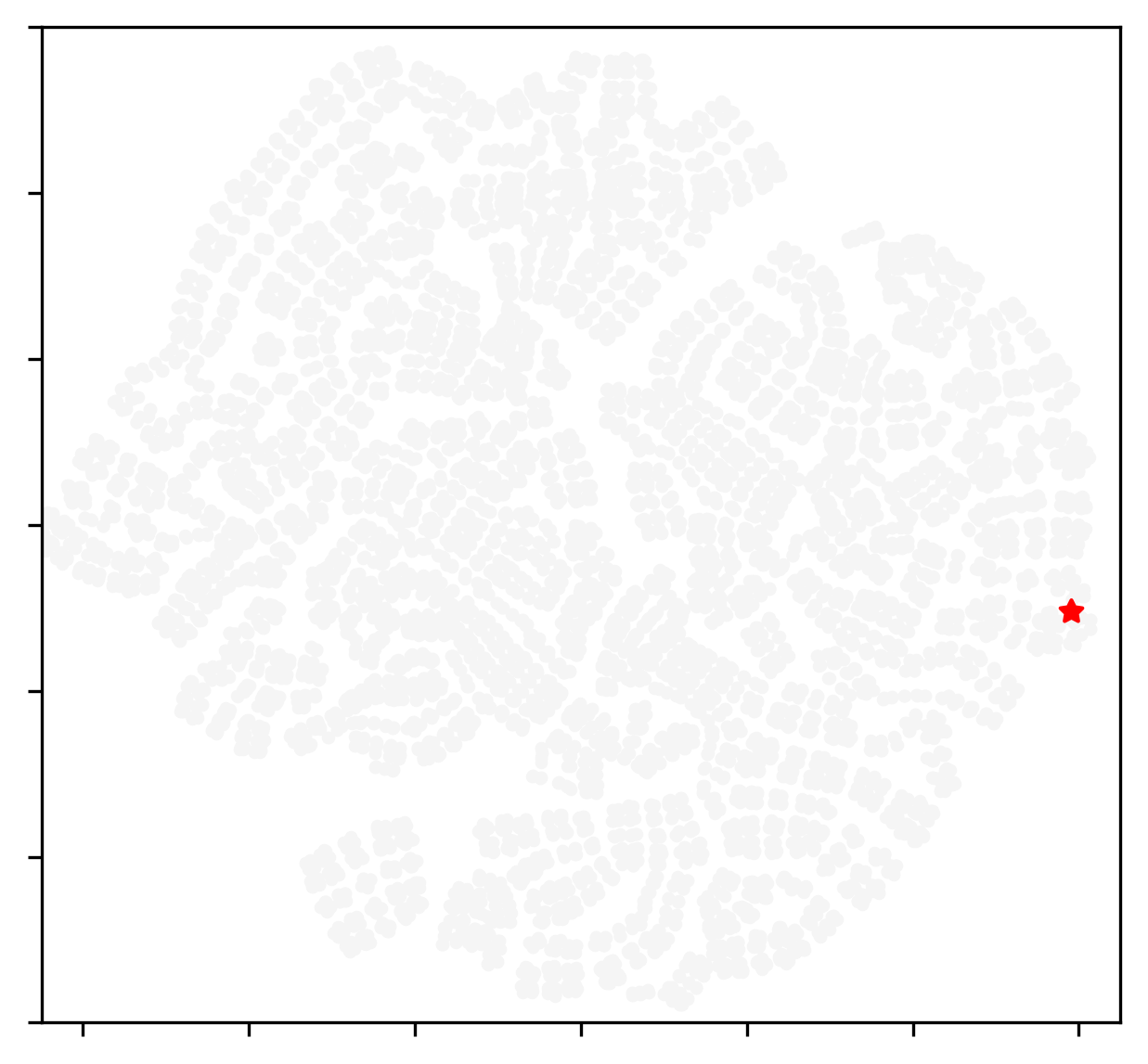}
	        \caption{The single candidate architecture found after the third stage. 
}
	        \label{fig:pruning3}
	    \end{subfigure}%
\end{minipage}
\caption{\textbf{TNAS hierarchically factorizes the search space and gradually prunes the unpromising architectures.} The colorbar shows the global rankings of architectures on CIFAR-10 \cite{krizhevsky2009cifar10} in NAS-Bench-201~\cite{Dong2020NASBench201ET}. Red stars indicate top-10 architectures. }
\vspace{-1em}
\label{fig:pruning}
\end{figure}

In this work, we diverge from the paradigms set by early NAS, and instead design a new algorithm to explore the search space in a wiser manner. Consider a search space $\mathcal{A}$ where the number of candidate operations is $M$ and the number of architecture layers to search is $L$. The size of the entire search space $|\mathcal{A}|$ equals to $M^L$. If $M=2$ or $L=1$, $|\mathcal{A}|$ can be drastically reduced to $2^L$ or $M$. The intuition behind our work is to develop a method that factorizes the operation space (size $M$) and the architecture layers (size $L$), and thus reduces the exploration size exponentially. 

\vspace{2pt}\noindent \textbf{Contributions.} \textbf{(1)}  We introduce an architecture tree and a binary operation tree to factorize the search space $L$ and $M$, respectively.  By combining the two trees, we iteratively branch a search space into two exclusive subspaces. \textbf{(2)} We propose a novel, flexible, accurate, and efficient NAS algorithm, called \textbf{TNAS}: NAS with trees. TNAS performs a modified bi-level Breadth-First Search (BFS) in the two proposed trees. By adjusting the expansion depths of the BFS, TNAS explicitly controls the exploration size $N$ and is able to exponentially reduce $N$ from $M^L$ to $O(L\log_2{M})$. The essence of TNAS is illustrated in Figure \ref{fig:pruning}. \textbf{(3)} TNAS is is able to find the global optimal architecture on CIFAR-10~\cite{krizhevsky2009cifar10} (94.37\% test acc.) in NAS-Bench-201~\cite{Dong2020NASBench201ET} within 4 GPU hours on one GTX2080Ti GPU. TNAS outperforms the RL and EA based NAS \cite{zoph2018nasnet,real2019regularized} as well as one-shot NAS \cite{pham2018efficient,chen2019pdarts}, with a similar search cost.

\begin{figure*}[t]
\centering
\begin{minipage}[b]{0.35\textwidth}
\centering
\begin{subfigure}{\textwidth}
	\includegraphics[width=\textwidth]{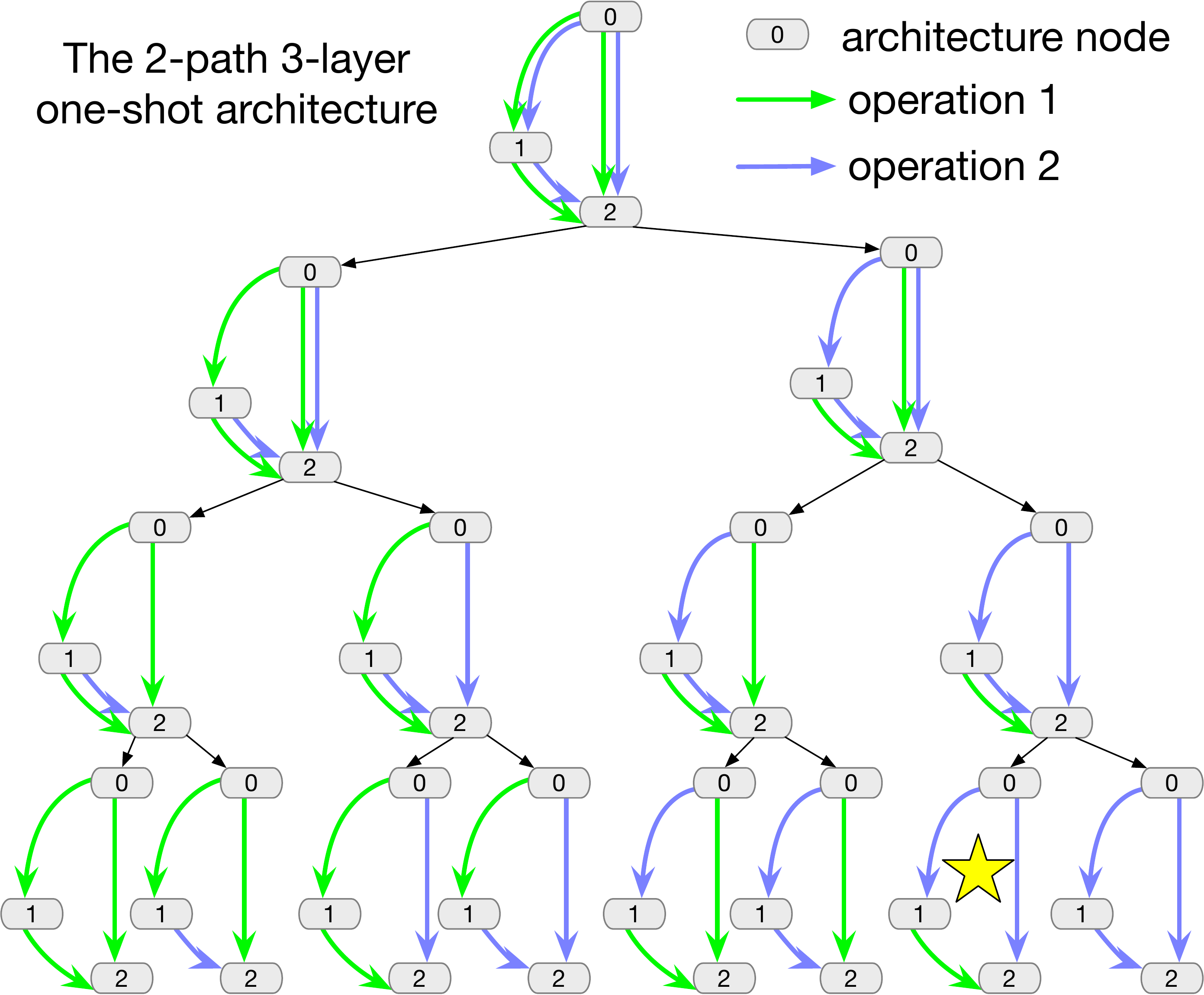}
    \caption{Architecture tree $\mathcal{T}_A$}
    \label{fig:arch_tree}
\end{subfigure}%
\end{minipage}
\hfill
\begin{minipage}[s]{0.63\textwidth}
\centering
\begin{subfigure}{\textwidth}
    \includegraphics[width=\linewidth]{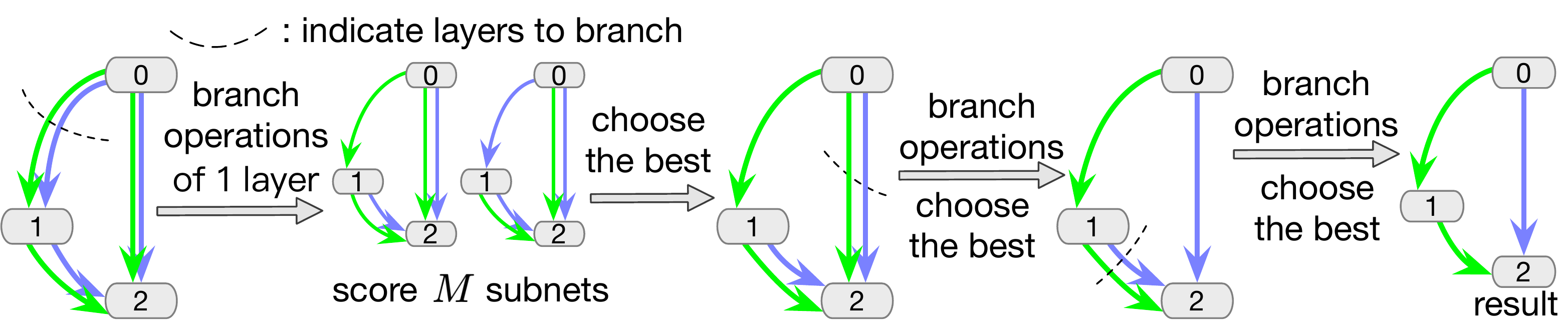}
    \caption{BFS with expansion depth $d_a=1$ explores $M\times L$ architectures}
    \label{fig:arch_tree_l1}
\end{subfigure}
\vfill
\begin{subfigure}{\textwidth}
    \includegraphics[width=\linewidth]{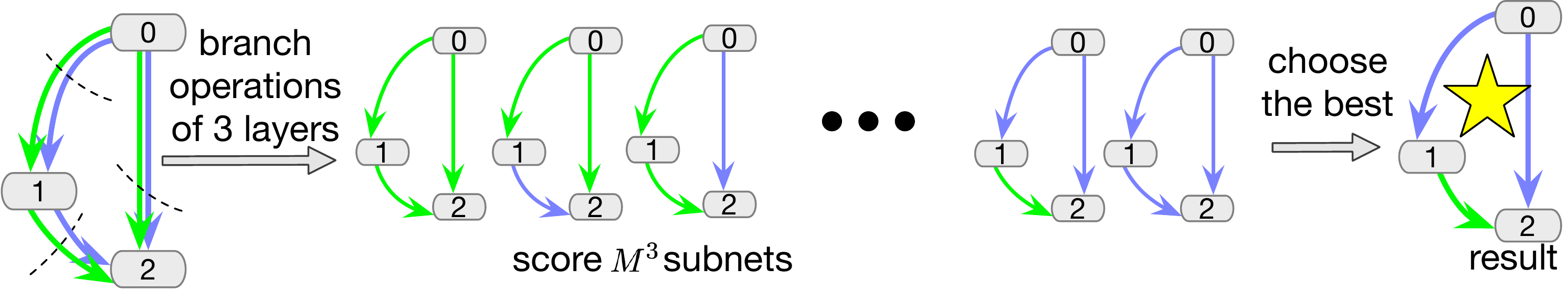}
    \caption{BFS with $d_a=L=3$ explores $M^L$ architectures}
    \label{fig:arch_tree_l3}
\end{subfigure}%
\end{minipage}
\caption{\textbf{Illustration of the architecture tree $\mathcal{T}_A$ and the proposed Breadth-First Search (BFS)}. 
}
\vspace{-1em}
\label{fig:arch_tree_pipelines}
\end{figure*}
\section{Related Work}\label{sec:related}
The computational bottleneck of NAS is \textit{exploring} candidate architectures in this huge search space and \textit{exploiting} each one (\ie score the architecture by training to convergence). Through the work, we name the number of architectures to score as the \textit{exploration size}, denoted as $N$. To alleviate the computational bottleneck, NAS algorithms should consider: (i) how to explore wisely, where time can be saved if the algorithm explores more among the ``good" architectures and less on the ``bad" ones, and  (ii) how to exploit wisely, where training each network to convergence just to know the architecture's performance then throwing weights away is inefficient. 

\mysection{Explore wisely.}
Early methods adopt Reinforcement Learning \cite{zoph17nas-rl,baker17metaqnn,zoph2018nasnet} or Evolutionary Algorithms \cite{Stanley2002EvolvingNN,Real17large_scale_evolution,real2019regularized} to auto-explore the huge search space. Although early NAS methods have been able to discover architectures that outperform manually designed networks, they consume significant computational resources. This is primarily because these algorithms require a large exploration size to achieve a decent search accuracy.
Progressive NAS is a method that factorizes the search space into a product of smaller search spaces and can greatly reduce the exploration size. PNAS~\cite{liu2018progressive} and P-DARTS~\cite{chen2019pdarts} start searching with shallow models and gradually progress to deeper ones. Li \etal propose block-wise progressive NAS~\cite{li2020dna,li2021bossnas} that consider the architectures is built by sequential blocks and search the architecture block by block.
SGAS~\cite{li2020sgas}, GreedyNAS~\cite{you2020greedynas}, and \cite{hu2020abs, zheng2020rethinking,wu2021weak} progressively shrink the search space by dropping unpromising candidates. These progressive NAS methods require a much smaller exploration size, but their greedy nature hampers their search accuracy. Our TNAS designs a new paradigm for exploring wisely by introducing two trees to factorize the search space.

\mysection{Exploit wisely.}
A straightforward idea of reducing exploitation is to train fewer epochs as done in Block-QNN~\cite{block_qnn}. A more advanced solution is to share weights among child networks, apart from training them from scratch. This weight-sharing strategy was first proposed by ENAS~\cite{brock2017smash,pham2018efficient} and has inspired many following works, including one-shot NAS~\cite{bender2018understanding,liu2018darts,guo2020spos,wan2020fbnet2,peng2020cream,yu2020bignas,chu2021darts-,gu2021dots}. To alleviate the evaluation degradation \cite{Yu20Evaluting,Bender2020CanWS,li2020sgas} issues of one-shot NAS caused by weight-sharing,  few-shot NAS~\cite{zhao2021fewshotnas,kshotnas} were proposed by training $k$ supernets instead of training only one. 
Another line of work to exploit wisely is accuracy prediction~\cite{liu2018progressive,chamnet,block_qnn_plus}, where an accuracy predictor is learned to directly estimate an architecture's accuracy without training it completely. 
Recently, metric-based NAS methods~\cite{mellor2020neural,chen2020tenas,zero-cost-nas,rlnas,hu2020abs,chen2021bnnas} have emerged, using well-designed metrics to score the sampled architectures quickly with significantly less training or even no training. 
Since our paper focuses on how to wisely explore the NAS search space, wise exploitation is an orthogonal direction. In fact, we highlight here that our TNAS can be applied with nearly all the aforementioned exploit-wise NAS methods. 

\section{Methodology} 
\label{sec:methodology}
We present TNAS (NAS with trees) to efficiently find a high-performance architecture by performing a modified bi-level Breadth-First Search in the proposed architecture tree $\mathcal{T}_A$ and binary operation tree $\mathcal{T}_O$. 

\subsection{Architecture Tree $\mathcal{T}_A$}
Given a search space with $L$ layers and $M$ operations per layer, we propose an architecture tree $\mathcal{T}_A$ to factorize the one-shot architecture and to exponentially reduce the exploration size. The architecture tree $\mathcal{T}_A$ is illustrated in \figLabel \ref{fig:arch_tree}. Each node in the tree represents an architecture. The root node is the $M$-path $L$-layer one-shot architecture. Each path in a layer denotes a distinct operation from $M$ candidate operations. 
$\mathcal{T}_A$ has a maximum depth level equal to $L$. For each node (architecture) at depth $i$ ($i \in [0, 1, \dots, L-1$), the tree separates the $M$ operations in layer $i$ into $M$ branches each with a single operation. Such branching is repeated for each node, until the leaf nodes are reached. Each leaf node represents a distinct single-path architecture. The union of the leaf nodes is the set of all candidate architectures. 
Note that if layer $i$ contains multiple operations, the output of this layer  will be the summation of the outputs of all operations at this layer, as inspired by the one-shot NAS \cite{bender2018understanding} and is formulated as: 
\begin{equation}\label{eqn:oneshot}
	{\bar{o}}^{i}(x) = \sum_{o_j \in \mathcal{O}} o_j^{i}(x),
\end{equation}
where $\mathcal{O} = \{o_j~|~ j = 1, 2, \ldots, M\}$ denotes $M$ different operations and $x$ denotes the input feature map.

\mysection{Breadth-First Search (BFS) in $\mathcal{T}_A$.}
Here, we show that the architecture search can be done by performing our modified BFS in the architecture tree $\mathcal{T}_A$. Our BFS requires a hyperparameter, the expansion depth denoted as $d_a$, where the subscript $a$ denotes ``architecture''. 
BFS starts at the root node (the one-shot model) at depth $0$, expands all its successors until depth $d_a$, and obtains up to $M^{d_a}$ leaf nodes after expansion. BFS scores the subnets defined by these leaf nodes, and picks the node with the highest score as the root node for the next step. The above procedure is defined as a \textit{decision step}, and is repeated until a single-path architecture is determined. The \textit{score} function can be chosen to be the validation performance after training, or  a metric function proposed by any metric-based NAS method such as the number of linear regions \cite{mellor2020neural}. For simplicity, we choose the scoring function to be validation performance in our experiments.
The expansion depth $d_a$ of the BFS denotes how many layers to branch at each decision step. 
As illustrated in \figLabel \ref{fig:arch_tree_l1}, the BFS with $d_a=1$ is a \textit{sequential, greedy} NAS algorithm that decides the operation for the architecture layer by layer, similar to the progressive NAS method SGAS \cite{li2020sgas}. 
The BFS with $d_a=L$ as shown in \figLabel \ref{fig:arch_tree_l3} works as the \textit{brute-force} NAS, where only $1$ decision step is required. The BFS explores all $M^{L}$ subnets and decides the operation for all of the layers at the same decision step. 
When $d_a=k\in \{2, \cdots, L-1\}$, $k$ layers are branched in each decision step, $M^k$ subnets need to be scored, and $\left\lceil{\frac{L}{k}}\right\rceil$ decision steps are required. This case works similar to the block-wise NAS \cite{li2020dna}, while our BFS does not require any block-level supervision. 


\subsection{Binary Operation Tree $\mathcal{T}_O$}
\begin{figure}[t]
\centering
\includegraphics[width=0.6\columnwidth]{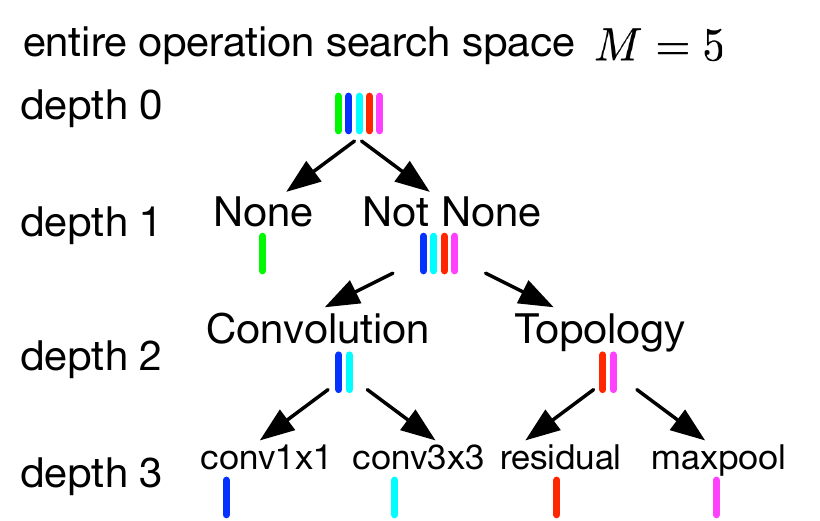}
\caption{\textbf{The binary operation tree $\mathcal{T}_O$}.
}
\vspace{-1em}
\label{fig:operation_tree}
\end{figure}

We propose a binary operation tree $\mathcal{T}_O$ that hierarchically factorizes the operation space to further reduce the exploration size. Each node in $\mathcal{T}_O$ is an \textit{operation group} consisting of one or more distinct operations. The root node represents $\mathcal{O}$, the entire operation space containing all $M$ operations. 
$\mathcal{T}_O$ starts from the root node and branches it into two child nodes that represent two exclusive operation groups. Such branching is repeated for each node until a leaf node that represents a single operation is reached. $\mathcal{T}_O$ has $M$ leaf nodes. The union of leaf nodes is $\mathcal{O}$. 
Taking NAS-Bench-201~\cite{Dong2020NASBench201ET} operation space as an example, we illustrate the $\mathcal{T}_O$ in \figLabel \ref{fig:operation_tree}.

\mysection{Breadth-First Search (BFS) in $\mathcal{T}_O$.}
The expansion depth of our modified BFS in $\mathcal{T}_O$ is denoted as $d_o$. BFS starts at the root node (the entire operation space) at depth $0$, expands all its successors until depth $d_o$, and obtains up to $2^{d_o}$ leaf nodes after expansion. These leaf nodes represent the current candidate operation groups. BFS scores the architectures equipped with these different operation groups, and picks the node defined by the operation group with the highest score as the root node for the next stage. 
The above procedure is defined as a \textit{decision stage}, and is repeated until a single operation is picked. 
Note that each architecture layer can choose different operation groups at each decision stage.
If $d_o = 1$, the BFS decides the operation groups per depth following $\mathcal{T}_O$. In this case, BFS consists of $\lceil\frac{\log_2{(M-1)}+1}{d_o}\rceil = 3$ decision stages. At the $1^{\text{st}}$ stage, BFS decides among \textit{None} or \textit{Not None} for each architecture layer. At the $2^{\text{nd}}$ stage, for those layers that chose \textit{Not None}, the algorithm decides among the \textit{Convolution} group or \textit{Topology} group. At the final stage, the algorithm will pick a single operation for each layer. 
If $d_o=3$, BFS only needs one decision stage to decide which single operation to choose for each layer.


\begin{figure}[t]
\centering
\includegraphics[width=0.50\textwidth]{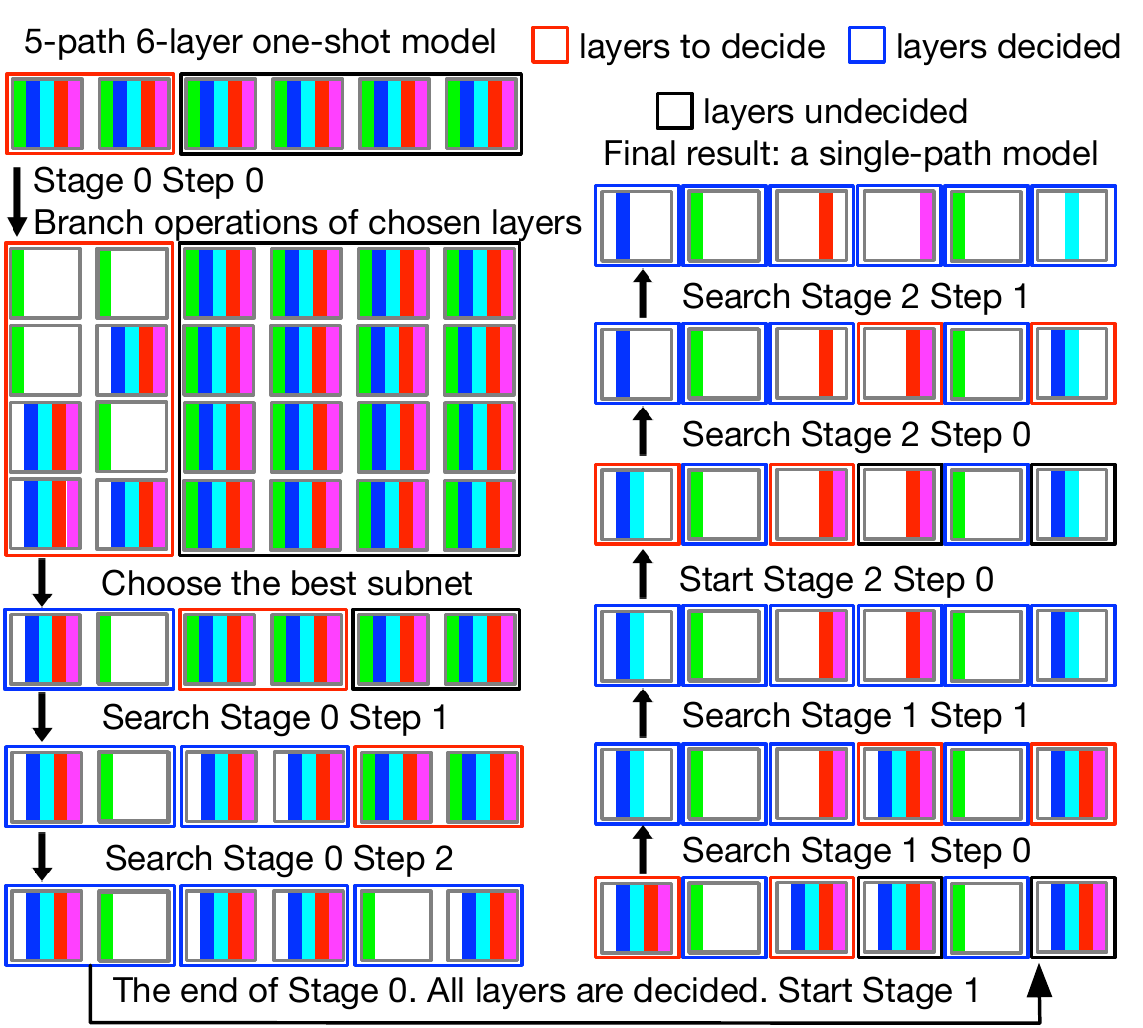}
\caption{\textbf{Illustration of TNAS ($d_a=2, d_o=1$).} 
}
\vspace{-1mm}
\label{fig:pipeline}
\end{figure}

\subsection{TNAS}\label{sec:tnas}
We present a new NAS algorithm:  \textbf{N}eural \textbf{A}rchitecture \textbf{S}earch with \textbf{T}rees (\textbf{TNAS}).
Given a search space with $M$ candidate operations and $L$ layers, TNAS constructs a binary operation tree $\mathcal{T}_O$ and an architecture tree $\mathcal{T}_A$.
TNAS starts from the $M$-path $L$-layer one-shot model, and performs \textit{bi-level Breadth-First Search} on $\mathcal{T}_O$ and $\mathcal{T}_A$. 
At the outer loop, TNAS performs BFS with the expansion depth $d_o=1$ on $\mathcal{T}_O$ by default, to make a large $d_a$ feasible. The outer loop requires $\lceil\log_2{(M-1)} +1\rceil$ decision stages. Each stage branches each operation group of the chosen layers into two child operation groups, which define the operation search space for the inner loop. The outer loop repeats the decision stage until every architecture layer reaches a leaf node of $\mathcal{T}_O$, \ie all the layers pick a single operation. In the inner loop, TNAS performs BFS with an expansion depth $d_a$ on $\mathcal{T}_A$. The inner loop takes $\left\lceil{\frac{L}{d_a}}\right\rceil$ decision steps. Each step chooses $d_a$ undecided layers to branch, obtains $2^{d_a}$ subnets, scores each subnet, and then chooses the highest scoring one. The chosen subnet will be used to replace the one-shot model and become the starting point for the next step. The inner loop repeats the above decision step until it chooses a leaf node of $\mathcal{T}_A$, \ie all layers of the architecture have decided their operation group at the current decision stage. We illustrate the TNAS algorithm ($d_o=1, d_a=2$) in \figLabel \ref{fig:pipeline}. The NAS-Bench-201 \cite{Dong2020NASBench201ET} search space (\ie $M=5$ and $L=6$) is used as an example.

%

\mysection{Exploration size analysis.} Given a search space with $M$ operations and $L$ layers, TNAS reduces the exploration size exponentially from $M^L$ to:  
\begin{equation}
\small
    N = O
     \left( 2^{d_od_a}  \times  \left\lceil\frac{L}{d_a}\right\rceil  \times \left\lceil\frac{\log_2{(M-1)}+1}{d_o}\right\rceil  \right)
\end{equation}
\section{Experiments}\label{sec:exp}

\begin{table}[t]
    \centering
    \caption{\textbf{State-of-the-art comparison on NAS-Bench-201}. Top-1 test accuracy (mean and standard deviation over 5 runs) are reported. For each dataset, \textbf{optimum} indicates the best test accuracy achievable in the NAS-Bench-201 search space.}
    \label{table:nasbench201}
    \vspace{-1em}
    \resizebox{.5\textwidth}{!}{
    \begin{tabular}{lcccccc}
    \toprule
    \textbf{Architecture} & \textbf{CIFAR-10} & \textbf{CIFAR-100} & \textbf{ImageNet-16-120} & \textbf{Search Cost (hours)} & \textbf{Search Method} \\ \midrule
    \textbf{optimum} & 94.37 & 73.51 & 47.31 & - & -\\ 
    \midrule
    ResNet \cite{he2016deep} & 93.97 & 70.86 & 43.63 & - & - \\ \midrule
    REA \cite{real2019regularized} & $93.92\pm0.30$ & $71.84\pm0.99$ & $45.54\pm1.03$ & 3.3 & EA \\
    REINFORCE \cite{ying2019nasbench101} & $93.85\pm0.37$ & $71.71\pm1.09$ & $45.24\pm1.18$ & 3.3 & RL \\
    RS \cite{bergstra2012random} & $93.70\pm0.36$ & $71.04\pm1.07$ & $44.57\pm1.25$ & 3.3 & random \\ \midrule
    NAS w.o. Training \cite{mellor2020neural} & $91.78\pm1.45$ & $67.05\pm2.89$ & $37.07\pm6.39$ & - & training-free \\
    TE-NAS \cite{chen2020tenas} & $93.90\pm0.47$ & $71.24\pm0.56$ & $42.38\pm0.46$ & - & training-free \\ \midrule
    RSPS \cite{li2020random} & $87.66\pm1.69$ & $58.33\pm4.34$ & $31.14\pm3.88$ & 2.2 & random \\
    ENAS \cite{pham2018efficient} & $54.30\pm0.00$ & $15.61\pm0.00$ & $16.32\pm0.00$ & 3.7 & EA \\
    DARTS (2nd) \cite{liu2018darts} & $54.30\pm0.00$ & $15.61\pm0.00$ & $16.32\pm0.00$ & 8.3 & gradient \\
    GDAS \cite{dong2019searching} & $93.61\pm0.09$ & $70.70\pm0.30$ & $41.84\pm0.90$ & 8.0 & gradient \\
    DARTS- \cite{chu2021darts-} & $93.80\pm0.40$ & $71.53\pm1.51$ & $45.12\pm0.82$ & 3.2 & gradient \\
    VIM-NAS \cite{wang2021vimnas} & $94.31\pm 0.11$ & $\textbf{73.07}\pm0.58$ & $46.27\pm0.17$ & - & gradient \\ 
    \midrule
	\textbf{TNAS (ours)} & \textbf{94.35}$\pm$0.03 & 73.02$\pm$0.34 & \textbf{46.31}$\pm$0.24& 3.6 &tree& \\
	\textbf{TNAS (best)} & $\textbf{94.37}$ & $73.09$ & \textbf{46.33} & 3.6 &tree& \\
    \bottomrule
    \end{tabular}
    }
\end{table}

\mysection{Setup.} 
We evaluate TNAS on NAS-Bench-201~\cite{Dong2020NASBench201ET} with $(d_o=1, d_a=6)$. We train each architecture over $2$ epochs and use the top-1 accuracy on validation set as the score for the architecture. If the architecture consists of a layer with multiple operations, the output of this layer is the sum of all outputs as Equation \ref{eqn:oneshot}. Note that other scoring methods aforementioned in Section \ref{sec:related} can also be applied.

\mysection{Results.}
Table \ref{table:nasbench201} compares TNAS with state-of-the-art NAS methods. \textit{TNAS finds the global optimal architecture in CIFAR-10 \cite{krizhevsky2009cifar10} within 4 GPU hours.} TNAS achieves 94.35\% average test accuracy, outperforming all other NAS methods.  We highlight that TNAS outperforms the EA (REA~\cite{real2019regularized}), RL (REINFORCE~\cite{ying2019nasbench101}) and random search (RS~\cite{bergstra2012random}) with a similar search cost, which clearly demonstrates the benefit of our NAS paradigm. TNAS also performs significantly better than the one-shot based methods, such as ENAS \cite{pham2018efficient}, DARTS\cite{liu2018darts}, GDAS \cite{dong2019searching} and DARTS- \cite{chu2021darts-}, while being more efficient.
TNAS picks one operation group out of two for each layer at each search stage, and thus prunes a portion of the candidate architectures in the search space, as illustrated in \figLabel \ref{fig:pruning}.

\section{Conclusion}\label{sec:conclusion}
We present a novel NAS algorithm, TNAS, that performs bi-level BFS on the proposed binary operation tree and the architecture tree. By adjusting the search depths on the trees, TNAS can explicitly control the exploration size. TNAS finds the global optimal architecture in NAS-Bench-201~\cite{Dong2020NASBench201ET} with a search cost of less than 4 GPU hours. 


\mysection{Acknowledgments} This work was done when Guocheng was remotely interned at Megvii technology. This work was also supported by the King Abdullah University of Science and Technology (KAUST) Office of Sponsored Research (OSR) through VCC funding.

{\small
\bibliographystyle{ieee_fullname}
\bibliography{tnas}

\begin{thebibliography}{10}\itemsep=-1pt

\bibitem{zero-cost-nas}
Mohamed~S. Abdelfattah, Abhinav Mehrotra, Lukasz Dudziak, and Nicholas~Donald
  Lane.
\newblock Zero-cost proxies for lightweight {NAS}.
\newblock In {\em International Conference on Learning Representations (ICLR)},
  2021.

\bibitem{baker17metaqnn}
Bowen Baker, Otkrist Gupta, Nikhil Naik, and Ramesh Raskar.
\newblock Designing neural network architectures using reinforcement learning.
\newblock In {\em {ICLR} (Poster)}. OpenReview.net, 2017.

\bibitem{bender2018understanding}
Gabriel Bender, Pieter-Jan Kindermans, Barret Zoph, Vijay Vasudevan, and Quoc
  Le.
\newblock Understanding and simplifying one-shot architecture search.
\newblock In {\em International Conference on Machine Learning}, pages
  549--558, 2018.

\bibitem{Bender2020CanWS}
Gabriel Bender, Hanxiao Liu, Bo Chen, Grace Chu, Shuyang Cheng, Pieter-Jan
  Kindermans, and Quoc~V. Le.
\newblock Can weight sharing outperform random architecture search? an
  investigation with tunas.
\newblock {\em 2020 IEEE/CVF Conference on Computer Vision and Pattern
  Recognition (CVPR)}, pages 14311--14320, 2020.

\bibitem{bergstra2012random}
James Bergstra and Yoshua Bengio.
\newblock Random search for hyper-parameter optimization.
\newblock {\em J. Mach. Learn. Res.}, 13:281--305, 2012.

\bibitem{brock2017smash}
Andrew Brock, Theodore Lim, James~M. Ritchie, and Nick Weston.
\newblock {SMASH:} one-shot model architecture search through hypernetworks.
\newblock In {\em {ICLR} (Poster)}. OpenReview.net, 2018.

\bibitem{chen2021bnnas}
Boyu Chen, Peixia Li, Baopu Li, Chen Lin, Chuming Li, Ming Sun, Junjie Yan, and
  Wanli Ouyang.
\newblock Bn-nas: Neural architecture search with batch normalization.
\newblock In {\em Proceedings of the IEEE/CVF International Conference on
  Computer Vision (ICCV)}, pages 307--316, October 2021.

\bibitem{chen2020tenas}
Wuyang Chen, Xinyu Gong, and Zhangyang Wang.
\newblock Neural architecture search on imagenet in four gpu hours: A
  theoretically inspired perspective.
\newblock In {\em International Conference on Learning Representations (ICLR)},
  2021.

\bibitem{chen2019pdarts}
Xin Chen, Lingxi Xie, Jun Wu, and Qi Tian.
\newblock Progressive differentiable architecture search: Bridging the depth
  gap between search and evaluation.
\newblock In {\em Proceedings of the IEEE/CVF International Conference on
  Computer Vision (ICCV)}, October 2019.

\bibitem{chu2021darts-}
Xiangxiang Chu, Xiaoxing Wang, Bo Zhang, Shun Lu, Xiaolin Wei, and Junchi Yan.
\newblock {DARTS-:} robustly stepping out of performance collapse without
  indicators.
\newblock In {\em {ICLR}}. OpenReview.net, 2021.

\bibitem{chamnet}
Xiaoliang Dai, Peizhao Zhang, Bichen Wu, Hongxu Yin, Fei Sun, Yanghan Wang,
  Marat Dukhan, Yunqing Hu, Yiming Wu, Yangqing Jia, Peter Vajda, Matt
  Uyttendaele, and Niraj~K. Jha.
\newblock Chamnet: Towards efficient network design through platform-aware
  model adaptation.
\newblock In {\em CVPR}, pages 11398--11407, 2019.

\bibitem{dong2019searching}
Xuanyi Dong and Yi Yang.
\newblock Searching for a robust neural architecture in four gpu hours.
\newblock In {\em Proceedings of the IEEE Conference on computer vision and
  pattern recognition}, pages 1761--1770, 2019.

\bibitem{Dong2020NASBench201ET}
Xuanyi Dong and Yi Yang.
\newblock Nas-bench-201: Extending the scope of reproducible neural
  architecture search.
\newblock In {\em {ICLR}}. OpenReview.net, 2020.

\bibitem{gu2021dots}
Yu-Chao Gu, Li-Juan Wang, Yun Liu, Yi Yang, Yu-Huan Wu, Shao-Ping Lu, and
  Ming-Ming Cheng.
\newblock Dots: Decoupling operation and topology in differentiable
  architecture search.
\newblock In {\em CVPR}, 2021.

\bibitem{guo2020spos}
Zichao Guo, X. Zhang, Haoyuan Mu, Wen Heng, Z. Liu, Y. Wei, and Jian Sun.
\newblock Single path one-shot neural architecture search with uniform
  sampling.
\newblock In {\em ECCV}, 2020.

\bibitem{he2016deep}
Kaiming He, Xiangyu Zhang, Shaoqing Ren, and Jian Sun.
\newblock Deep residual learning for image recognition.
\newblock In {\em Proceedings of the IEEE conference on computer vision and
  pattern recognition}, pages 770--778, 2016.

\bibitem{hu2020abs}
Yiming Hu, Yuding Liang, Zichao Guo, Ruosi Wan, X. Zhang, Yichen Wei, Qingyi
  Gu, and Jian Sun.
\newblock Angle-based search space shrinking for neural architecture search.
\newblock {\em ArXiv}, abs/2004.13431, 2020.

\bibitem{krizhevsky2009cifar10}
Alex Krizhevsky.
\newblock Learning multiple layers of features from tiny images.
\newblock 2009.

\bibitem{li2020dna}
Changlin Li, Jiefeng Peng, Liuchun Yuan, Guangrun Wang, Xiaodan Liang, Liang
  Lin, and Xiaojun Chang.
\newblock Block-wisely supervised neural architecture search with knowledge
  distillation.
\newblock In {\em Proceedings of the IEEE/CVF Conference on Computer Vision and
  Pattern Recognition (CVPR)}, June 2020.

\bibitem{li2021bossnas}
Changlin Li, Tao Tang, Guangrun Wang, Jiefeng Peng, Bing Wang, Xiaodan Liang,
  and Xiaojun Chang.
\newblock {B}oss{NAS}: Exploring hybrid {CNN}-transformers with block-wisely
  self-supervised neural architecture search.
\newblock In {\em ICCV}, 2021.

\bibitem{li2020sgas}
Guohao Li, Guocheng Qian, Itzel~C. Delgadillo, Matthias M{\"{u}}ller, Ali~K.
  Thabet, and Bernard Ghanem.
\newblock {SGAS:} sequential greedy architecture search.
\newblock In {\em {CVPR}}, pages 1617--1627. {IEEE}, 2020.

\bibitem{li2020random}
Liam Li and Ameet Talwalkar.
\newblock Random search and reproducibility for neural architecture search.
\newblock In {\em Uncertainty in Artificial Intelligence}, pages 367--377.
  PMLR, 2020.

\bibitem{liu2018progressive}
Chenxi Liu, Barret Zoph, Maxim Neumann, Jonathon Shlens, Wei Hua, Li-Jia Li, Li
  Fei-Fei, Alan Yuille, Jonathan Huang, and Kevin Murphy.
\newblock Progressive neural architecture search.
\newblock In {\em Proceedings of the European Conference on Computer Vision
  (ECCV)}, pages 19--34, 2018.

\bibitem{liu2018darts}
Hanxiao Liu, Karen Simonyan, and Yiming Yang.
\newblock Darts: Differentiable architecture search.
\newblock In {\em {ICLR} (Poster)}. OpenReview.net, 2019.

\bibitem{mellor2020neural}
Joe Mellor, Jack Turner, Amos~J. Storkey, and Elliot~J. Crowley.
\newblock Neural architecture search without training.
\newblock In {\em {ICML}}, volume 139 of {\em Proceedings of Machine Learning
  Research}, pages 7588--7598. {PMLR}, 2021.

\bibitem{peng2020cream}
Houwen Peng, Hao Du, Hongyuan Yu, Qi Li, Jing Liao, and Jianlong Fu.
\newblock Cream of the crop: Distilling prioritized paths for one-shot neural
  architecture search.
\newblock In {\em NeurIPS}, 2020.

\bibitem{pham2018efficient}
Hieu Pham, Melody~Y. Guan, Barret Zoph, Quoc~V. Le, and Jeff Dean.
\newblock Efficient neural architecture search via parameter sharing.
\newblock In {\em {ICML}}, volume~80 of {\em Proceedings of Machine Learning
  Research}, pages 4092--4101. {PMLR}, 2018.

\bibitem{real2019regularized}
Esteban Real, Alok Aggarwal, Yanping Huang, and Quoc~V Le.
\newblock Regularized evolution for image classifier architecture search.
\newblock In {\em Proceedings of the AAAI Conference on Artificial
  Intelligence}, volume~33, pages 4780--4789, 2019.

\bibitem{Real17large_scale_evolution}
Esteban Real, Sherry Moore, Andrew Selle, Saurabh Saxena, Yutaka~Leon Suematsu,
  Jie Tan, Quoc~V. Le, and Alexey Kurakin.
\newblock Large-scale evolution of image classifiers.
\newblock In {\em {ICML}}, volume~70 of {\em Proceedings of Machine Learning
  Research}, pages 2902--2911. {PMLR}, 2017.

\bibitem{Stanley2002EvolvingNN}
K. Stanley and R. Miikkulainen.
\newblock Evolving neural networks through augmenting topologies.
\newblock {\em Evolutionary Computation}, 10:99--127, 2002.

\bibitem{kshotnas}
Xiu Su, Shan You, Mingkai Zheng, Fei Wang, Chen Qian, Changshui Zhang, and
  Chang Xu.
\newblock K-shot {NAS:} learnable weight-sharing for {NAS} with k-shot
  supernets.
\newblock In Marina Meila and Tong Zhang, editors, {\em ICML}, volume 139,
  pages 9880--9890, 2021.

\bibitem{wan2020fbnet2}
Alvin Wan, Xiaoliang Dai, Peizhao Zhang, Zijian He, Yuandong Tian, Saining Xie,
  Bichen Wu, Matthew Yu, Tao Xu, Kan Chen, Peter Vajda, and Joseph~E. Gonzalez.
\newblock Fbnetv2: Differentiable neural architecture search for spatial and
  channel dimensions.
\newblock In {\em {CVPR}}, pages 12962--12971. Computer Vision Foundation /
  {IEEE}, 2020.

\bibitem{wang2021vimnas}
Yaoming Wang, Yuchen Liu, Wenrui Dai, Chenglin Li, Junni Zou, and Hongkai
  Xiong.
\newblock Learning latent architectural distribution in differentiable neural
  architecture search via variational information maximization.
\newblock In {\em {ICCV}}, pages 12292--12301. {IEEE}, 2021.

\bibitem{wu2021weak}
Junru Wu, Xiyang Dai, Dongdong Chen, Yinpeng Chen, Mengchen Liu, Ye Yu,
  Zhangyang Wang, Zicheng Liu, Mei Chen, and Lu Yuan.
\newblock Stronger nas with weaker predictors.
\newblock {\em arXiv preprint arXiv:2102.10490}, 2021.

\bibitem{ying2019nasbench101}
Chris Ying, Aaron Klein, Esteban Real, Eric Christiansen, Kevin~P. Murphy, and
  Frank Hutter.
\newblock Nas-bench-101: Towards reproducible neural architecture search.
\newblock In {\em ICML}, 2019.

\bibitem{you2020greedynas}
Shan You, Tao Huang, Mingmin Yang, Fei Wang, Chen Qian, and Changshui Zhang.
\newblock Greedynas: Towards fast one-shot nas with greedy supernet.
\newblock In {\em Proceedings of the IEEE/CVF Conference on Computer Vision and
  Pattern Recognition}, pages 1999--2008, 2020.

\bibitem{yu2020bignas}
Jiahui Yu, Pengchong Jin, Hanxiao Liu, Gabriel Bender, Pieter{-}Jan Kindermans,
  Mingxing Tan, Thomas~S. Huang, Xiaodan Song, Ruoming Pang, and Quoc Le.
\newblock Bignas: Scaling up neural architecture search with big single-stage
  models.
\newblock In {\em {ECCV} {(7)}}, volume 12352 of {\em Lecture Notes in Computer
  Science}, pages 702--717. Springer, 2020.

\bibitem{Yu20Evaluting}
Kaicheng Yu, Christian Sciuto, Martin Jaggi, Claudiu Musat, and Mathieu
  Salzmann.
\newblock Evaluating the search phase of neural architecture search.
\newblock In {\em {ICLR}}. OpenReview.net, 2020.

\bibitem{rlnas}
Xuanyang Zhang, Pengfei Hou, Xiangyu Zhang, and Jian Sun.
\newblock Neural architecture search with random labels.
\newblock In {\em The IEEE Conference on Computer Vision and Pattern
  Recognition (CVPR)}, pages 10907--10916, 2021.

\bibitem{zhao2021fewshotnas}
Yiyang Zhao, Linnan Wang, Yuandong Tian, Rodrigo Fonseca, and Tian Guo.
\newblock Few-shot neural architecture search.
\newblock In {\em {ICML}}, volume 139 of {\em Proceedings of Machine Learning
  Research}, pages 12707--12718. {PMLR}, 2021.

\bibitem{zheng2020rethinking}
Xiawu Zheng, Rongrong Ji, Qiang Wang, Qixiang Ye, Zhenguo Li, Yonghong Tian,
  and Qi Tian.
\newblock Rethinking performance estimation in neural architecture search.
\newblock {\em 2020 IEEE/CVF Conference on Computer Vision and Pattern
  Recognition (CVPR)}, pages 11353--11362, 2020.

\bibitem{block_qnn}
Zhao Zhong, Junjie Yan, Wei Wu, Jing Shao, and Cheng{-}Lin Liu.
\newblock Practical block-wise neural network architecture generation.
\newblock In {\em The IEEE Conference on Computer Vision and Pattern
  Recognition (CVPR)}, pages 2423--2432, 2018.

\bibitem{block_qnn_plus}
Zhao Zhong, Zichen Yang, Boyang Deng, Junjie Yan, Wei Wu, Jing Shao, and
  Cheng{-}Lin Liu.
\newblock Blockqnn: Efficient block-wise neural network architecture
  generation.
\newblock {\em {IEEE} Trans. Pattern Anal. Mach. Intell.}, 43(7):2314--2328,
  2021.

\bibitem{zoph17nas-rl}
Barret Zoph and Quoc~V. Le.
\newblock Neural architecture search with reinforcement learning.
\newblock In {\em {ICLR}}. OpenReview.net, 2017.

\bibitem{zoph2018nasnet}
Barret Zoph, Vijay Vasudevan, Jonathon Shlens, and Quoc~V Le.
\newblock Learning transferable architectures for scalable image recognition.
\newblock In {\em Proceedings of the IEEE conference on computer vision and
  pattern recognition (CVPR)}, pages 8697--8710, 2018.

\end{thebibliography}
}

\end{document}